\documentclass[letterpaper]{article} 
\usepackage{aaai23}  
\usepackage{times}  
\usepackage{helvet}  
\usepackage{courier}  
\usepackage[hyphens]{url}  
\usepackage{graphicx} 
\usepackage{natbib}  
\usepackage{caption} 
\usepackage{algorithm}
\usepackage{algorithmic}

\usepackage{multirow}
\usepackage{booktabs}

\usepackage{amssymb}
\usepackage{graphicx}
\usepackage{amsmath}

\newcommand\scalemath[2]{\scalebox{#1}{\mbox{\ensuremath{\displaystyle #2}}}}

\usepackage{fontawesome}
\usepackage{pgfplots}
\usetikzlibrary{positioning,matrix,calc,arrows.meta,decorations.pathreplacing,shapes.geometric}
\definecolor{ao(english)}{rgb}{0.0, 0.5, 0.0}
\definecolor{alizarin}{rgb}{0.82, 0.1, 0.26}

\usepackage{newfloat}
\usepackage{listings}
\DeclareCaptionStyle{ruled}{labelfont=normalfont,labelsep=colon,strut=off} 
\floatstyle{ruled}
\newfloat{listing}{tb}{lst}{}
\floatname{listing}{Listing}
\usepackage{hyperref}

\title{ESRL: Efficient Sampling-based Reinforcement Learning \\for Sequence Generation}
\author{
    Chenglong Wang\textsuperscript{\rm 1},
    Hang Zhou\textsuperscript{\rm 1},
    Yimin Hu\textsuperscript{\rm 1},
    Yifu Huo\textsuperscript{\rm 1}, 
    Bei Li\textsuperscript{\rm 1}, \\
    Tongran Liu\textsuperscript{\rm 3},
    Tong Xiao\textsuperscript{\rm 1,2}\footnote{Corresponding author.} and
    Jingbo Zhu\textsuperscript{\rm 1,2}
}
\affiliations{
    \textsuperscript{\rm 1} School of Computer Science and Engineering, Northeastern University, Shenyang, China \\
    \textsuperscript{\rm 2} NiuTrans Research, Shenyang, China \\
    \textsuperscript{\rm 3} CAS Key Laboratory of Behavioral Science, Institute of Psychology, CAS, Beijing, China \\
    {\{clwang1119, ctrl.hang\}@gmail.com},
    {\{xiaotong, zhujingbo\}@mail.neu.edu.cn}
}

\usepackage{bibentry}

\begin{document}

\maketitle

\begin{abstract}
Applying Reinforcement Learning (RL) to sequence generation models enables the direct optimization of long-term rewards (\textit{e.g.,} BLEU and human feedback), but typically requires large-scale sampling over a space of action sequences.
This is a computational challenge as presented by the practice of sequence generation problems, such as machine translation, where we often deal with a large action space (\textit{e.g.,} a vocabulary) and a long action sequence (\textit{e.g.,} a translation).
In this work, we introduce two-stage sampling and dynamic sampling approaches to improve the sampling efficiency during training sequence generation models via RL.
We experiment with our approaches on the traditional sequence generation tasks, including machine translation and abstractive summarization.
Furthermore, we evaluate our approaches in RL from human feedback (RLHF) through training a large language model using the reward model.  
Experimental results show that the efficient sampling-based RL, referred to as ESRL, can outperform all baselines in terms of both training efficiency and memory consumption.
Notably, ESRL yields consistent performance gains over the strong REINFORCE, minimum risk training, and proximal policy optimization methods.
\end{abstract}

\section{Introduction}
The use of Reinforcement Learning (RL) in training sequence generation models has gained significant attention in recent years.
This is primarily due to the fact that sequence generation is inherently a long-term decision-making problem and RL is particularly well-suited for optimizing long-term rewards, such as sequence-level scores \cite{wieting2019beyond, donato2022mad} and human  feedbacks \cite{nguyen2017reinforcement, stiennon2020learning,  ouyang2022training, chatgpt2022}.
Additionally, by leveraging an autoregressive mode of generation, RL training can significantly mitigate the \textit{exposure bias} problem \cite{ranzato2015sequence, wang2020exposure}.

The RL training process typically involves two steps: (1) sampling a number of candidate sequences with a pre-trained model given an input (call it \textit{exploration}), and (2) using an RL method, such as REINFORCE \cite{williams1992simple} and Proximal Policy Optimization (PPO) \cite{schulman2017proximal}, to optimize the model with the long-term rewards given the sampled sequences (call it \textit{optimization}). 
This paradigm has achieved promising results on several sequence generation tasks, such as machine translation \cite{wieting2019beyond, yehudai2022reinforcement, donato2022mad}, abstractive summarization \cite{celikyilmaz2018deep, stiennon2020learning}, and dialogue generation \cite{hsueh2020semantic}.
Moreover, it has been proved to have a promising potential for guiding a large language model (LLM) to learn from human feedbacks \cite{ouyang2022training, chatgpt2022}.  

Despite such successes, applying RL to NLP is not low-hanging fruit. In practical applications of sequence generation, we often deal with a large action space (\textit{e.g.,} a vocabulary) and a long action sequence (\textit{e.g.,} a translation). This poses a serious computational challenge to the exploration procedure \cite{keneshloo2019deep}, and is an important factor motivating the design of sophisticated sampling approaches.

To mitigate this problem, we investigate strategies for reducing the computational burden of exploration when applying RL to sequence generation models.
In this work, we propose an \textbf{E}fficient \textbf{S}ampling-based \textbf{RL} (ESRL) method, which enables more efficient exploration by using the following two approaches.
For one, we use a two-stage sampling framework to implement the exploration.
It can take full advantage of the Transformer's parallelism computation, so the excessive computational graph storage requirements disappear.
Furthermore, we propose a dynamic sampling approach that can reduce redundant sampling by considering the capability of a model.
The motivation is that heavy sampling is simply not necessary because pre-trained generation models have already acquired some ability of generation.

We experiment with the proposed ESRL on machine translation and summarization tasks based on Transformer \cite{vaswani2017attention}.
Experimental results show that ESRL can surpass both the REINFORCE \cite{williams1992simple, kiegeland2021revisiting} and minimum risk training \cite{shen2016minimum} in terms of generation quality, training time, and memory consumption.
Notably, compared to REINFORCE, it can reduce 47\% of the memory consumption and 39\% of the training time on the machine translation task.
Additionally, our ESRL significantly outperforms the vanilla Transformer over 1.04 BLEU scores on the IWSLT’14 De-En and WMT’14 En-De test sets.
It also significantly outperforms all baselines on the abstractive summarization task.
Furthermore, we evaluate our ESRL in RLHF \cite{christiano2017deep} with LLaMA-7B-LoRA \cite{hu2021lora, touvron2023llama}.
The results demonstrate that ESRL remains significantly more memory-efficient and faster in RLHF while achieving an improvement of +30.00 points on Vicuna’s total score, as evaluated by GPT-4 \cite{vicuna2023}, compared to the robust PPO \cite{schulman2017proximal}.

\section{Related Work}
While reinforcement learning (RL) has long been appreciated in robotics and other fields, it has recently emerged as a promising approach to advance sequence generation models \cite{ranzato2015sequence, celikyilmaz2018deep, yehudai2022reinforcement, donato2022mad}.
For example, \citet{edunov2018classical} compared objective functions commonly used in RL for sequence generation models.
\citet{choshen2019weaknesses} and \citet{kiegeland2021revisiting} examined the limitations of RL in neural machine translation.
Moreover, \citet{kiegeland2021revisiting} conducted experiments on in-domain and cross-domain adaptation setups to highlight the significance of exploration during RL training.
It is also an upward trend in using RL to train large language models with human preferences \cite{nguyen2017reinforcement, stiennon2020learning,  ouyang2022training, chatgpt2022}.

As another line of research, researchers focused on exploring better reward functions to enhance the learning of generation models, such as the use of semantic similarity \cite{li2016deep, wieting2019beyond, yasui2019using} and the design of learned reward functions \cite{shi2018toward, bohm2019better, shu2021reward}.
More recent work aimed at addressing the challenge of large action spaces in sequence generation models \cite{hashimoto2018accelerated, yehudai2022reinforcement}.

Although previous work improves the performance of RL on sequence generation tasks, they are often hindered by the inefficient exploration problem.
Researchers have been aware of this \cite{keneshloo2019deep}, but it is still rare to see studies on this issue.

\section{Our Method}
In this section, we first recall the preliminaries of using RL in training sequence generation models. 
Then, we present our efficient sampling-based RL method. 
Last, we introduce our optimization algorithm.

\subsection{Preliminaries}
\label{sec_preliminaries}
\paragraph{Sequence Generation Model}
Given an input $x$ such as a text, a sequence generation model generates a sequence of $N$ tokens $y = \{y_{1},\dots, y_{N}\}$, where each token $y_{t}$ is drawn from a vocabulary.
At the training stage, the model learns the probability:
\begin{eqnarray}
    p_{\theta}(y|x) = \prod_{t=1}^{N}p_{\theta}(y_{t}|y_{<t},x)
\label{eq-generation-pro}
\end{eqnarray}
where $y_{<t}$ is the prefix $\left\lbrace y_{1}, y_{2}, \dots, y_{t-1}\right\rbrace $, and $\theta$ is a set of model parameters.
In this process, the standard training objective is to maximize the likelihood over all the tokens of the target sequence, \textit{i.e., maximum likelihood estimation (MLE)} \cite{myung2003tutorial}.
At the inference stage, we generate tokens sequentially according to probability $p_{\theta}$.
In this paper, we consider the tasks of neural machine translation and abstractive summarization and use them as instances of the above model.

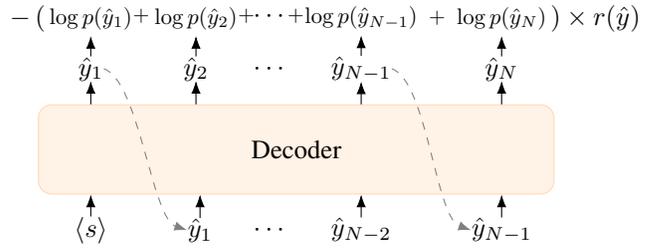
\begin{figure}[t!]
	\centering
	\tikzstyle{every node}=[scale=1]
	\begin{tikzpicture}
		\node [draw=orange!30,fill=orange!10,minimum width=19.5em, minimum height=7.5ex,rounded corners=5pt] at (0,0) (decoder) {Decoder};

		\node 	[inner sep=0pt, anchor=center,]	at 	([xshift=2em, yshift=-3ex]decoder.south west) 	(s below) 		{\textit{$\left\langle s \right\rangle$}};
		\node 	[inner sep=0pt, anchor=center,] at 	([xshift=-2em, yshift=-3ex]decoder.south east) 	(yn-1 below) 	{$\hat{y}_{N-1}$};
		\node 	[inner sep=0pt, anchor=west,] 	at 	([xshift=3em]s below.east) 						(y1 below) 		{$\hat{y}_1$};
		\node 	[inner sep=0pt, anchor=east,] 	at 	([xshift=-3em]yn-1 below.west) 					(yn-2 below) 	{$\hat{y}_{N-2}$};

		\node	[inner sep=0pt, anchor=center,]	at ([xshift=2em, yshift=3ex]decoder.north west) 	(y1 above) 		{$\hat{y}_1$};
		\node	[inner sep=0pt, anchor=center,]	at ([xshift=-2em, yshift=3ex]decoder.north east) 	(yn above) 		{$\hat{y}_{N}$};
		\node	[inner sep=0pt, anchor=west,]	at ([xshift=3em]y1 above.east) 						(y2 above) 		{$\hat{y}_2$};
		\node	[inner sep=0pt, anchor=center,]	at (yn-2 below.north|-yn above.east) 				(yn-1 above) 	{$\hat{y}_{N-1}$};
		
		\node 	[inner sep=0pt, anchor=center, scale=0.85]	at	([yshift=3.2ex]y1 above.north)                  (log p y1) 		{$\log p(\hat{y}_1)$};
		\node 	[inner sep=0pt, anchor=center, scale=0.85]	at	([yshift=3.2ex]y2 above.north)                  (log p y2) 		{$\log p(\hat{y}_2)$};
		\node 	[inner sep=0pt, anchor=center, scale=0.85]	at	([yshift=3.2ex]yn-1 above.north)                (log p yn-1) 	{$\log p(\hat{y}_{N-1})$};
		\node 	[inner sep=0pt, anchor=center, scale=0.85]	at	([yshift=3.2ex]yn above.north)                  (log p yn) 		{$\log p(\hat{y}_{N})$};
		\node 	[inner sep=0pt, anchor=east, scale=1]		at 	([xshift=0em]log p y1.west) 						            {$-\left( \right.$};
		\node 	[inner sep=0pt, anchor=west, scale=1]		at 	([xshift=0em]log p yn.east) 						            {$\left. \right)\times r(\hat{y})$};
		\node 	[inner sep=0pt, anchor=west, scale=0.8]		at	([yshift=.2ex]log p y1.east)						            {\textbf{$+$}};
		\node 	[inner sep=0pt, anchor=west, scale=0.8]		at	([yshift=.2ex]log p y2.east)                    (plus y2) 		{\textbf{$+$}};
		\node 	[inner sep=0pt, anchor=east, scale=0.8]		at	([yshift=.2ex]log p yn-1.west)                  (plus yn-1)		{\textbf{$+$}};
		\node 	[inner sep=0pt, anchor=center, scale=0.8]	at	($1/2*(log p yn-1.east)+1/2*(log p yn.west)$)		            {\textbf{$+$}};
		\node 												at 	($1/2*(plus y2)+1/2*(plus yn-1)$)               (log p dots)	{$\cdots$};
		\node 	[inner sep=0pt, anchor=center,]				at 	(log p dots|-y2 above.east)                     (dots below) 	{$\cdots$};
		\node	[inner sep=0pt, anchor=center,]				at 	(log p dots|-y1 below.east)                     (dots above) 	{$\cdots$};

		\draw 	[-Latex]	(y1 above.south|-decoder.north)		--	(y1 above.south);
		\draw 	[-Latex]	(y2 above.south|-decoder.north)		--	(y2 above.south);
		\draw 	[-Latex]	(yn-1 above.south|-decoder.north) 	--	(yn-1 above.south);
		\draw 	[-Latex]	(yn above.south|-decoder.north) 	--	(yn above.south);

		\draw 	[-Latex]	(y1 above.north)      --	([yshift=-0.5ex]log p y1.south);
		\draw 	[-Latex]	(y2 above.north)      --	([yshift=-0.5ex]log p y2.south);
		\draw 	[-Latex]	(yn-1 above.north)    --	([yshift=-0.5ex]log p yn-1.south);
		\draw 	[-Latex]	(yn above.north)      --	([yshift=-0.5ex]log p yn.south);
		
		\draw 	[-Latex]	(y1 below.north) 					--	(y1 below.north|-decoder.south);
		\draw 	[-Latex]	(s below.north) 					--	(s below.north|-decoder.south);
		\draw 	[-Latex]	(yn-2 below.north) 					--	(yn-2 below.north|-decoder.south);
		\draw 	[-Latex]	(yn-1 below.north) 					--	(yn-1 below.north|-decoder.south);


		\draw	[-Latex, dashed, draw=gray]	(y1 above.east)		..	controls([xshift=1.35em]y1 above.south east)	and	([xshift=-1.35em]y1 below.north west)	..	(y1 below.west);
		\draw	[-Latex, dashed, draw=gray]	(yn-1 above.east)	..	controls([xshift=1.35em]yn-1 above.south east)	and	([xshift=-1.35em]yn-1 below.north west)	..	(yn-1 below.west);
		
	\end{tikzpicture}
	\def\red cross
		{
			\begin{tikzpicture}
				\draw[red,pos=.5,rotate=45,scale=1.4,thick] (0,2 pt) -- (0,-2 pt) ;
				\draw[red,pos=.5,rotate=-45,scale=1.4,thick] (0,2 pt) -- (0,-2 pt) ;
			\end{tikzpicture}
		}
        \vspace{-3mm}
	\caption{
		  An illustration of the traditional RL loss calculation.
            In this process, we need to store each of computational graphs produced by $\left\{p(\hat{y}_{1}), p(\hat{y}_{2}), \cdots, p(\hat{y}_{N})\right\}$ to calculate the gradients.
            Thus, the memory footprint grows drastically as the sampled sequence become longer.
	}
        \vspace{-6mm}
	\label{fig-vanilla-rl}
\end{figure} 

\paragraph{Long-term Reward Optimization}
Given a pre-trained sequence generation model, we can use RL to train this model. RL seeks to maximize the long-term reward, written as $\arg\max_{\theta}\mathbb{E}p_{\theta}(\hat{y}|x)[r(\hat{y})]$, where $\left< x, y\right> $ is a training instance, $\hat{y}$ is a generated sequence, and $r(\cdot)$ is a reward function computing the long-term reward for $\hat{y}$.
$r(\cdot)$ is typically defined to be a standard metric function, such as BLEU \cite{papineni2002bleu} and ROUGE \cite{lin2004rouge}.
The corresponding RL loss for this training instance is then given by:
\begin{eqnarray}
    \mathcal{L}_{\mathrm{RL}} = \sum_{\hat{y} \in \Omega(x)}p_{\theta}(\hat{y}|x) r(\hat{y})
    \label{eq-Lrl}
\end{eqnarray}
where $\Omega(x)$ is the output space which comprises all possible candidate target sequences for input $x$.

\begin{figure*}[t!]
	\centering
        \tikzstyle{every node}=[scale=1.05]
	\begin{tikzpicture}

		\tikzstyle{back blk} = [draw=gray!30,rounded corners=4pt,minimum height=11em,anchor=west];
		\tikzset{text of back blk/.style={right=0em of #1.north west,anchor=north west}}
		\node [back blk,fill=red!2,minimum width=22.5em] at (0,0) (b1) {};
		\node [text of back blk=b1,scale=0.9,yshift=-1ex] (b1_text) {\small Dynamic Sampling
		};
		\node [back blk,fill=blue!2,minimum width=24em] at ([xshift=.8em]b1.east) (b2) {};
		\node [text of back blk=b2,scale=0.9,yshift=-1ex] {\small Two-stage Sampling};
		\node at ($1/2*(b1.south west)+1/2*(b1_text.south west)$) (b1 west center) {};
		
		\matrix
		[matrix of nodes,
		nodes in empty cells,
		row sep={1.2em,between origins},
		matrix anchor=west,
		nodes={anchor=center,inner sep=1pt},
		right=1em of b1 west center.east,
		below=4ex of b1 west center,
		inner sep=0pt,
		xshift=.2em] (vec input)
		{
			\scriptsize $[$ $x_1$ $x_2$ $\cdots$ $x_n$ $]$\\
		};
		\node[below=0.2ex of vec input,
		align=center,
		inner sep=0em,
		scale=0.85]
		{\scriptsize Input Batch \par};

		\tikzstyle{enc doc block} = [minimum width=3.2em,minimum height=4em,rounded corners=3pt,fill=#1!10,draw=#1!20,inner sep=1pt,anchor=center];
		\node [enc doc block=blue] at ([xshift=0em,yshift=4ex]b1 west center-|vec input) (enc) {\scriptsize Encoder};
		\node [below=0.3ex of enc.north,anchor=north,inner sep=0pt,label={[inner sep=0pt,yshift=-.2ex,scale=0.8]below:{\textcolor{ao(english)}{\scriptsize w/ gradient}}},scale=0.8] {\textcolor{ao(english)}{\scriptsize \faLink}};
		
		\tikzstyle{c} = [draw,dashed,rounded corners=2pt,minimum width=1.75em,minimum height=8.5ex];
		\node [c,right=1em of enc.east,anchor=west] (enc out) {};
		\tikzstyle{enc out ele} = [minimum height=1.5ex,minimum width=1.5em,rounded corners=2pt,inner sep=0pt];
		\node [enc out ele,above=2.5ex of enc out.center,fill=red!30] (c1) {\scriptsize $c_1$};
		\node [enc out ele,above=.5ex of enc out.center,fill=orange!30] (c2) {\scriptsize $c_2$};
		\node [enc out ele,below=2.5ex of enc out.center,fill=blue!30] (cn) {\scriptsize $c_n$};
		\node at ($1/2*(c2.south)+1/2*(cn.north)$) [inner sep=0pt,scale=0.6] {\small \rotatebox{90}{$\cdots$}};
		
		\node [shape=diamond,right=1em of enc out,draw,minimum width=3.6em,minimum height=4.8ex,fill=red!10,draw=red!20,align=center] (condition selection) {};
		\node at (condition selection.center) [yshift=-0.2ex,scale=0.6,align=center,execute at begin node=\setlength{\baselineskip}{7pt}] {\scriptsize Check Model \\ \scriptsize Capability};

		\node [enc doc block=orange,anchor=west] at ([xshift=3.7em]condition selection.east|-enc.west) (wo dec) {\scriptsize Decoder};
        \node at ([yshift=-1ex,xshift=0em]wo dec.south) [scale=0.8] {\scriptsize Greedy Search};
		\node [below=0.3ex of wo dec.north,anchor=north,inner sep=0pt,label={[inner sep=0pt,yshift=-.2ex,scale=0.8]below:{\textcolor{alizarin}{\scriptsize w/o gradient}}},scale=0.8] {\textcolor{alizarin}{\scriptsize \faUnlink}};
		
		
		capacity
		\matrix
		[matrix of nodes,
		nodes in empty cells,
		row sep={1.2em,between origins},
		matrix anchor=west,
		every node/.style={anchor=center,inner sep=1pt,scale=.6},
		scale=.6,
		right=1.2em of wo dec.east|-b1 west center,
		inner sep=0pt,
		above delimiter=[,
		below delimiter={]}] (vec cap)
		{
			$Cap_{x_1}$\\
			$Cap_{x_2}$\\
			\node [anchor=center,inner sep=0pt] {\rotatebox{-90}{$\cdots$}};\\
			$Cap_{x_n}$\\
		};
		
		\tikzstyle{ssb} = [inner sep=1pt, outer sep=0pt,minimum height=.8em,minimum width=.8em,fill=#1!30,rounded corners=1pt];
		\node [ssb=orange,anchor=south west] at ([xshift=1.1em,yshift=0ex]b1 west center-|b2.west) (ssb2) {\scriptsize 0};
		\node [ssb=red,anchor=south] at (ssb2.north) (ssb1) {\scriptsize 2};
		\node [ssb=gray,anchor=north,label={[scale=.5]center:\rotatebox{90}{$\cdots$}}] at (ssb2.south) (ssbd) {};
		\node [ssb=blue,anchor=north] at (ssbd.south) (ssbd) {\scriptsize 4};
		\node
		[below=1.5ex of ssbd,
		align=center,
		inner sep=0em,
		execute at begin node=\setlength{\baselineskip}{1.6ex},
		scale=0.7]
		{\scriptsize Sampling \\ \scriptsize Size \par};
		
		\tikzstyle{rsb} = [draw,dashed,rounded corners=2pt,minimum width=1.75em,minimum height=15.5ex];
		\node [rsb,right=1em of ssb2.south east,anchor=west] (rsb) {};
		\tikzstyle{rsb ele} = [minimum height=1.5ex,minimum width=1.5em,rounded corners=2pt,inner sep=0pt,anchor=center];
		\node [rsb ele,above=6ex of rsb.center,fill=red!30] {\scriptsize $c_1$};
		\node [rsb ele,above=4ex of rsb.center,fill=red!30] (rsbc1) {\scriptsize $c_1$};
		\node [rsb ele,below=0ex of rsb.center,fill=blue!30] (rsbcn) {\scriptsize $c_n$};
		\node [rsb ele,below=2ex of rsb.center,fill=blue!30] {\scriptsize $c_n$};
		\node [rsb ele,below=4ex of rsb.center,fill=blue!30] {\scriptsize $c_n$};
		\node [rsb ele,below=6ex of rsb.center,fill=blue!30] {\scriptsize $c_n$};
		\node at ($1/2*(rsbc1.south)+1/2*(rsbcn.north)$) [inner sep=0pt,scale=0.9] {\scriptsize \rotatebox{90}{$\cdots$}};
		\node at ([yshift=-1ex,xshift=1em]{{$1/2*(ssb2.south)+1/2*(rsb)$}|-rsb.south}) [scale=0.95] {\scriptsize Restructuring Batch};
		
		\node [enc doc block=orange,anchor=west] at ([xshift=1.2em]rsb.east) (dec) {\scriptsize Decoder};
        \node at ([yshift=-1ex,xshift=0em]dec.south) [scale=0.8] {\scriptsize Autoregressive Mode};
		\node [below=0.3ex of dec.north,anchor=north,inner sep=0pt,label={[inner sep=0pt,yshift=-.2ex,scale=0.8]below:{\textcolor{alizarin}{\scriptsize w/o gradient}}},scale=0.8] {\textcolor{alizarin}{\scriptsize \faUnlink}};
		
		\node [scale=.6,execute at begin node=\setlength{\baselineskip}{2ex},align=center,anchor=center] at ([yshift=0.5ex]dec.north|-rsb.north) (adj temp) {\small Adjusted \\ \small Temperature};
		
		\tikzstyle{transb} = [draw,dashed,rounded corners=2pt,minimum width=1.75em,minimum height=15.5ex];
		\node [transb,right=2.5em of dec.east,anchor=west] (trans) {};
		\tikzstyle{trans ele} = [minimum height=1.5ex,minimum width=1.5em,rounded corners=2pt,inner sep=0pt,anchor=center];
		\node [trans ele,above=6ex of trans.center,fill=red!30] (trans 11) {\scriptsize $\hat{y}_{11}$};
		\node [trans ele,above=4ex of trans.center,fill=red!30] (trans 12) {\scriptsize $\hat{y}_{12}$};
		\node [trans ele,below=0ex of trans.center,,fill=blue!30] at (trans.center) (trans n1) {\scriptsize $\hat{y}_{n1}$};
		\node [trans ele,below=2ex of trans.center,fill=blue!30] {\scriptsize $\hat{y}_{n2}$};
		\node [trans ele,below=4ex of trans.center,fill=blue!30] {\scriptsize $\hat{y}_{n3}$};
		\node [trans ele,below=6ex of trans.center,fill=blue!30] (trans n4) {\scriptsize $\hat{y}_{n4}$};
		\node at ($1/2*(trans 12)+1/2*(trans n1)$) [inner sep=0pt,scale=0.9] {\scriptsize \rotatebox{90}{$\cdots$}};
		\node at ([yshift=-1ex]trans.south) [scale=0.95] {\scriptsize Sampled Sequences};

		\node [enc doc block=orange,anchor=west] at ([xshift=1.2em]trans.east) (dec 2) {\scriptsize Decoder};
        \node at ([yshift=-1ex,xshift=0em]dec 2.south) [scale=0.8] {\scriptsize Parallelism};
		\node [below=0.3ex of dec 2.north,anchor=north,inner sep=0pt,label={[inner sep=0pt,yshift=-.2ex,scale=0.8]below:{\textcolor{ao(english)}{\scriptsize w/ gradient}}},scale=0.8] {\textcolor{ao(english)}{\scriptsize \faLink}};
		
		\tikzstyle{prob b} = [draw,dashed,rounded corners=2pt,minimum width=4.3em,minimum height=15.5ex];
		\node [prob b,right=1.2em of dec 2.east,anchor=west] (probs) {};
		\tikzstyle{probs ele} = [minimum height=1.5ex,minimum width=4em,rounded corners=2pt,inner sep=0pt,anchor=center];
		\node [probs ele,above=6ex of probs.center,fill=red!30] (probs 11) {\scriptsize $p_\theta(\hat{y}_{11}|x_1)$};
		\node [probs ele,above=4ex of probs.center,fill=red!30] (probs 12) {\scriptsize $p_\theta(\hat{y}_{12}|x_1)$};
		\node [probs ele,below=0ex of probs.center,,fill=blue!30] at (probs.center) (probs n1) {\scriptsize $p_\theta(\hat{y}_{n1}|x_n)$};
		\node [probs ele,below=2ex of probs.center,fill=blue!30] {\scriptsize $p_\theta(\hat{y}_{n2}|x_n)$};
		\node [probs ele,below=4ex of probs.center,fill=blue!30] {\scriptsize $p_\theta(\hat{y}_{n3}|x_n)$};
		\node [probs ele,below=6ex of probs.center,fill=blue!30] (probs n4) {\scriptsize $p_\theta(\hat{y}_{n4}|x_n)$};
		\node at ($1/2*(probs 12)+1/2*(probs n1)$) [inner sep=0pt,scale=0.9] {\scriptsize \rotatebox{90}{$\cdots$}};
		\node at ([yshift=-1ex]probs.south) [scale=0.95] {\scriptsize Probabilities};

		
		\tikzstyle{above the arrow} = [midway,align=center,above,execute at begin node=\setlength{\baselineskip}{1ex},scale=.8]
		\tikzstyle{below the arrow} = [midway,align=center,below,execute at begin node=\setlength{\baselineskip}{1ex},scale=.8]
		\draw [-Latex] (vec input) -- (enc);
		\draw [-Latex] (enc) -- (enc out);
		\draw [-Latex] (enc out) -- (condition selection);
		\draw [-Latex] (condition selection) -- (wo dec.west) node [above the arrow,scale=.85] {\small Unrecorded};
		\draw [-Latex,rounded corners=1.5pt] (condition selection.south) -- ({$2*(vec cap.west)-(wo dec.east)$}-|condition selection.south) -- ([xshift=-.7em]vec cap.west|-{$2*(vec cap.west)-(wo dec.east)$}) node [below the arrow,scale=.85,xshift=0em] {\small Recorded} -- ([xshift=-.7em]vec cap.west) -- (vec cap.west);
		\draw [-Latex,rounded corners=1.5pt] (wo dec) -- ([xshift=-.7em]wo dec.east-|vec cap.west) -- ([xshift=-.7em]vec cap.west) -- (vec cap.west);
		\draw [-Latex] (vec cap.east) -- ([xshift=-.1em]ssb2.south west);
		\draw [-Latex] ([xshift=.1em]ssb2.south east) -- (rsb);
		\draw [-Latex] (rsb) -- (dec);
		\draw [-Latex,rounded corners=1.5pt] (ssb1.north) -- (ssb1.north|-adj temp.east) -- (adj temp);
		\draw [-Latex,rounded corners=1.5pt] ([xshift=-.3em]adj temp.east) -- ({$1/2*(dec.east)+1/2*(trans n1.west)$}|-adj temp.east) -- ([yshift=0em]ssb1.north-|{$1/2*(dec.east)+1/2*(trans n1.west)$});
		\draw [-Latex] (dec.east) -- (trans.west) node [above the arrow,scale=.8] {\small Sampling};
		\draw [-Latex] (trans) -- (dec 2);
		\draw [-Latex] (dec 2) -- (probs);
	\end{tikzpicture}
	\caption{
		  Architecture of ESRL.
            We introduce two-stage sampling and dynamic sampling approaches to design ESRL, which enables it to be much more memory-efficient and much faster in training a sequence generation model.
            For the two-stage sampling, we take full advantage of Transformer’s parallelism computation to avoid the excessive computational graph storage.
            During the dynamic sampling, based on the estimated capability, we dynamically adjust the size and temperature of sampling to eliminate unnecessary exploration.
            Here encoder portion is used only by the encoder-decoder sequence generation model.
	}
        \vspace{-4mm}
	\label{fig_main_image}
\end{figure*}
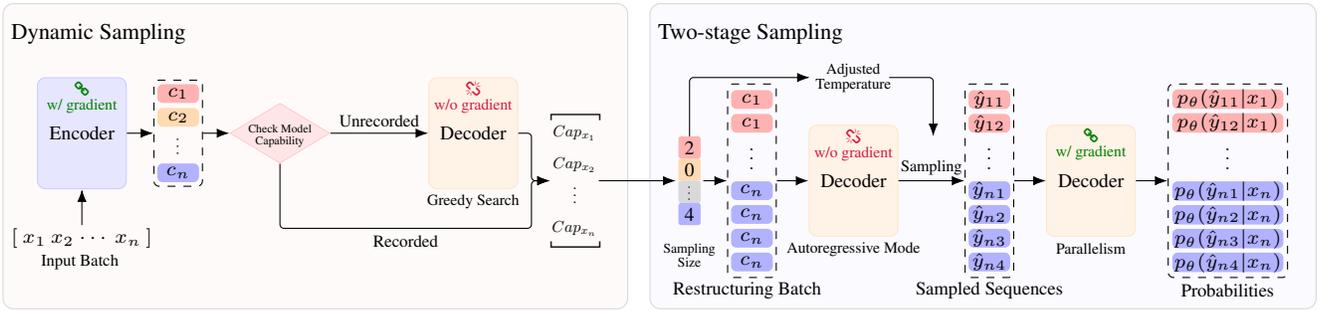

\paragraph{Exploration}
However, computing Eq. \ref{eq-Lrl} is intractable because the size of $\Omega(x)$ grows exponentially with the size of the vocabulary and the lengths of the target sequences.
To address this challenge, RL usually performs exploration to approximate $\Omega(x)$.
A commonly-used method to solve Eq. \ref{eq-Lrl} is the Monte Carlo method \cite{williams1992simple}.
For each training instance, a number of sequences are sampled from a multinomial distribution defined by a Softmax layer with a temperature factor \cite{choshen2019weaknesses}.
Here, both the sampling size and the sampling temperature can be used to control to what extent we explore the space.
For example, a larger scale means more candidates involved in sampling, and a higher temperature means a larger diversity of sampled sequences \cite{kiegeland2021revisiting}.

\paragraph{Policy Gradient}
To optimize the model with the long-term rewards of sampled sequences, policy gradient methods, such as REINFORCE \cite{williams1992simple} and minimum risk training (MRT) \cite{shen2016minimum}, are often used.
Specifically, REINFORCE uses log derivatives to define the loss function:
\begin{eqnarray}
    \mathcal{L}_{\mathrm{REINFORCE}} =  -\sum_{\hat{y} \in S(x)}\log p_{\theta}(\hat{y}|x) r(\hat{y})
    \label{eq-pg}
\end{eqnarray}
where $S(x)$ is an approximated space, which consists of these sampled sequences.
The calculation process is also illustrated in Figure \ref{fig-vanilla-rl}.
Since each sequence is sampled by an autoregressive mode, RL training needs to store more computational graphs (approximately $N$ times) for each training instance than MLE. 

Unlike REINFORCE, MRT method uses these sampled sequences to approximate a posterior distribution with renormalization:
\begin{eqnarray}
    Q_{\theta}(\hat{y}|x) = \frac{p_{\theta}(\hat{y}|x)^{\alpha}}{\sum_{\hat{y}\in S(x)}p_{\theta}(\hat{y}|x)^{\alpha}}
    \label{eq-mrt}
\end{eqnarray}
where $\alpha$ is a smoothness parameter and $Q_{\theta}(\hat{y}|x)$ is a distribution defined on the approximated space.
Based on the $Q_{\theta}$ distribution, MRT gives a new loss function:
\begin{eqnarray}
    \mathcal{L}_{\mathrm{MRT}} = \sum_{\hat{y}\in S(x)}Q_{\theta}(\hat{y}|x)[-r(\hat{y})]
\end{eqnarray}
In some cases, MRT can achieve better performance compared with REINFORCE \cite{kiegeland2021revisiting}.
But the exploration process of MRT requires an enormous amount of memory to store the computational graphs used in renormalization.

\subsection{Efficient Sampling-based RL (ESRL)}
In this work, our aim is to reduce the computational cost of applying RL to sequence generation models.
We propose the ESRL to achieve this.
The overview of ESRL is depicted in Figure \ref{fig_main_image}.
As shown in the figure, we present two-stage sampling and dynamic sampling in ESRL achieve our goal.
In the following subsections, we will describe them in detail.

\subsubsection{Two-stage Sampling}
In response to excessive computational graph storage requirements produced by the sampling process, we use a two-stage framework that effectively mitigates this issue.
Stage one is to sample the candidate sequences with an autoregressive mode.
Note that this stage is not involved in backpropagation.
It thus does not require the storage of computational graphs.
Stage two is to calculate the probabilities of the sampled candidate sequences. \textit{i.e.,} $p_{\theta}(\hat{y}|x)$ in Eqs. \ref{eq-pg} and \ref{eq-mrt}.
At this stage, due to the presence of the complete output sequence, we can use Transformer's parallelism computation instead of an autoregressive mode.
It allows this calculation to be done with just one forward pass.
Compared to the conventional sampling approach, the two-stage sampling approach incurs additional time costs due to an extra forward pass. 
However, with the help of two-stage sampling, it can effectively reduce the memory footprint.
Generally, the conventional sampling of RL training needs to store the computational graphs of $N$ forward passes, while the two-stage sampling only stores the computational graph of one forward pass. 

\paragraph{Dynamic Sampling}
We propose a dynamic sampling approach to further improve the efficiency of RL training.
In our dynamic sampling approach, we first estimate the model capability, then adjust the sampling size and temperature according to this estimated capability so that we can perform sampling in an adequate and efficient way.

For the model capability estimation, we reuse old sequences sampled at the previous epoch.
Specifically, given an input $x$, following the sampling of candidate sequences, we employ these sampled sequences to estimate the model's generation capability of the input.
Then, the estimated model capability is then recorded and used in the subsequent epoch to adjust the sampling size for the same input.
Taking the machine translation task as an instance, we use the entropy \cite{settles2009active} and BLEU \cite{papineni2002bleu} to estimate the model capability.
When using BLEU to estimate the model capability, the capability score is given by:
\begin{eqnarray}
    Cap_{x} = \frac{1}{m} \sum_{\hat{y}\in S(x)}\mathrm{BLEU}(\hat{y}, y)
\end{eqnarray}
where $m$ is the sampling size of the input $x$ and $\mathrm{BLEU}(\cdot)$ is the \textit{sacre}BLEU \cite{post2018call}.
When considering entropy as another estimation of the model capability, written as:
\begin{eqnarray}
    \begin{aligned}
        &Cap_{x} = \\ &1+\frac{1}{N\times m}\sum_{\hat{y}\in S(x)}\sum_{t=1}^{N}p_{\theta}(\hat{y}_{t}|\hat{y}_{<t},x)\log p_{\theta}(\hat{y}_{t}|\hat{y}_{<t},x)
    \end{aligned}
\end{eqnarray}
There are other choices to define $Cap_{x}$ for specific tasks.
For instance, we can replace BLEU with ROUGE \cite{lin2004rouge} in the abstractive summarization task.
Note that when the model's capability for a given input $x$ is not recorded, \textit{i.e.,} no sampling operation has been performed on the input, we employ a greedy search algorithm to generate an optimal sequence quickly.
Then we use this generated sequence to estimate the model capability. 

For the sampling size adjustment, our main aim is to eliminate unnecessary exploration.
Specifically, when $Cap_{x}$ is high, we consider that the model has ability to get a great long-term reward and thus decrease the sampling size.
By contrast, when $Cap_{x}$ is low, we increase the sampling size to have a larger-scale exploration.
It allows the model to learn from a sufficient number of possible generated sequences per input and to improve its own capability.
Here we use the following function to achieve this goal:
\begin{eqnarray}
    k_{x} = \lceil k_{max} -\beta \cdot n \cdot k_{max} \cdot \frac{Cap_{x}}{\sum_{x\in I}Cap_{x}} \rceil
\end{eqnarray}
where $k_{x}$ and $k_{max}$ denote the adjusted sampling size and the maximum sampling size, respectively. 
$\beta$ is a ratio of eliminated samples within the range of [0, 1), relative to the total number of samples.
Here we use batch-level elimination strategy that reduces the sampling size of the input with higher capability score within the current batch's distribution.
Thus $I$ denotes an input set consisting of all inputs in the current batch and $n$ denotes the number of inputs.

Considering that the sampling temperature also impacts the exploration, we adopt a simple strategy to control the exploration by adjusting the temperature: when the capacity score is low, we use a higher temperature to encourage exploration.
We dynamically adjust the temperature in the interval $[\tau_{min}, \tau_{max}]$ based on the adjusted sampling size to further control the exploration.
The rule of temperature adjustment is given by:
\begin{eqnarray}
	\tau_{x} = \tau_{min}+k_{x}\times \frac{\tau_{max}-\tau_{min}}{k_{max}}
\end{eqnarray}
where $\tau_{x}$ is the adjusted temperature for $x$.

After adjusting the size and temperature of sampling, we sample $k_{x}$ candidate sequences for each input.
Following \citet{kiegeland2021revisiting}'s work, we use a restructuring batch trick which restructures a new batch by repeating the encoder representations to act as the input of the decoder (see Figure \ref{fig_main_image}), to take advantage of the parallel computation.

\subsection{Optimization}
We replace the standard policy method with the fusion of MRT and REINFORCE in computing the loss.
Specifically, we use $\mathcal{L}_{\mathrm{MRT}}$ to serve as the loss when $k_{x}>1$.
When $k_{x}=1$, since the renormalization is not feasible, we instead use $\mathcal{L}_{\mathrm{REINFORCE}}$ to serve as the loss.
This design combines the strengths of MRT and REINFORCE and makes full use of the sampled sequences to optimize the model.

\paragraph{FIFO-based Baseline Reward}
The \textit{baseline reward} technique \cite{sutton2018reinforcement} has been shown to be effective to improve the generalization of sequence generation models \cite{kreutzer2017bandit}.
The ideal baseline value is an average of the long-term rewards of all possible candidate sequences.
Again, this is intractable because there is an exponentially large number of candidate sequences in sequence generation tasks.
Although some works attempt to estimate this ideal baseline value \cite{hashimoto2018accelerated}, they involve complex training.
Inspired by the idea of using a queue to proxy the global in \citet{wang2021selective}'s work, we propose a FIFO-based baseline reward approach, which employs a First-In-First-Out (FIFO) global reward queue $\mathcal{Q}$ to compute the baseline value.
We use $\mathcal{Q}_{size}$ to denote the reward queue size. 
At each training step, we push rewards of all sampled sequences into $\mathcal{Q}$ and pop out the ‘Oldest’ rewards.
Then we compute the average of the rewards in $\mathcal{Q}$ to serve as the baseline value $b$.
By using this baseline reward, we replace the reward function in Eqs. \ref{eq-pg} and \ref{eq-mrt} with $r(\hat{y},y)-b$.

\section{Experiments}
We evaluated our ESRL method on the traditional sequence generation tasks, including machine translation and abstractive summarization.
We also evaluated ESRL in RLHF with LLaMA-7B-LoRA.
\subsection{Experimental Setups}
\paragraph{Datasets} 
The datasets used for each task are as follows:
\begin{itemize}
    \item \textit{Machine Translation}: 	
    We conducted experiments on two machine translation datasets, including a small-scale IWSLT’14 German-English (De-En) dataset and a large-scale WMT’14 English-German (En-De) dataset.
    We preprocessed the datasets following the same setup in \citet{hu2021ranknas}'s work.
    \item \textit{Abstractive Summarization}: 
    We also tested the ESRL's capability to train the abstractive summarization model on the CNN/DM dataset \cite{hermann2015teaching}.
    Our data preprocess method was the same as in \citet{li2022learning}.
    \item \textit{RLHF}:
    We employed Alpaca’s 52k data and GPT-4 Alpaca data (English) \cite{peng2023instruction, taori2023stanford} to perform the supervised fine-tuning (SFT) and RLHF.
    We used the GPT-4 Comparison English dataset\footnote{\url{https://github.com/Instruction-Tuning-with-GPT-4/GPT-4-LLM}} to train our reward model.
\end{itemize}

\begin{table*}[ht]
    \centering
    \scalebox{0.88}{
    \begin{tabular}{lccccrrcccrr}
    \toprule[1.1pt]
    \multirow{2}{*}{Method} & \multirow{2}{*}{SS} & \multicolumn{5}{c}{IWSLT'14 De-En}  & \multicolumn{5}{c}{WMT'14 En-De}  \\ \cmidrule(l){3-7} \cmidrule(l){8-12} 
 &    & SBLEU & BLEU & COMET-22 & \begin{tabular}[c]{@{}c@{}}Time\\ (hours)\end{tabular} & \begin{tabular}[c]{@{}c@{}}Memory\\ (G)\end{tabular} & SBLEU &BLEU & COMET-22 & \begin{tabular}[c]{@{}c@{}}Time\\ (hours)\end{tabular} & \begin{tabular}[c]{@{}c@{}}Memory\\ (G)\end{tabular} \\ \midrule
    MLE            & - &33.77 & 34.57  & 79.32  &  -     &  - &26.73  & 27.26    &  83.36       &  -    &  - \\ \midrule
    REINFORCE     & 1  &33.91      &34.73          &79.52      &4.52      &3.31      &26.97        &27.45     &83.45         &7.70 &5.32  \\
    ESRL-Random   & 1  &33.71      &34.52          &79.47      &2.69      &2.63      &26.85        &27.32     &83.38        &6.45 &4.84  \\
    ESRL-BLEU     & 1  &\bf34.02   &\bf34.84       &\bf79.62   &3.15      &\bf2.54   &\bf27.02     &\bf27.51  &\bf83.55      &\bf6.19 &4.55  \\
    ESRL-Entropy  & 1  &33.96      &34.80          &79.56      &\bf2.73   &2.66      &26.95     &27.43  &83.43        &6.26 &\bf4.13 \\  \midrule
    REINFORCE     & 5  &34.05      &34.87          &79.58      &7.04      &8.66      &27.10     &27.53     &83.52       &12.68 &13.83 \\
    MRT           & 5  &34.17      &34.91          &79.66      &8.96      &16.60     &27.12     &27.63     &83.60       &13.76 &14.93  \\
    ESRL-Random   & 5  &33.68      &34.45          &79.23      &\bf3.73   &5.26      &26.87     &27.36     &83.41        &10.34 &12.39  \\
    ESRL-BLEU     & 5  &34.37      &\bf35.16       &79.81      &4.34      &5.49      &\bf27.25     &\bf27.74  &\bf83.68       &11.14 &11.79 \\
    ESRL-Entropy  & 5  &\bf34.40   &35.14          &\bf79.85   &3.99      &\bf5.19   &27.18        &27.62     &83.59        &\bf10.29 &\bf11.75  \\  \midrule
    REINFORCE     & 10  &34.22      &34.96          &79.63      &8.19     &15.61     &27.21     &27.65     &83.72    &15.20 &20.85  \\
    MRT           & 10  &34.31      &35.01          &79.71      &10.26    &23.03     &27.26     &27.75     &83.80    &16.90 &22.14  \\
    ESRL-Random   & 10  &33.75      &34.54          &79.31      &\bf4.63  &10.15     &27.05     &27.48     &83.52      &13.26 &15.87  \\
    ESRL-BLEU     & 10  &34.53      &35.37     &80.02  &4.85      &\bf9.86           &\bf27.43  &27.94     &83.87      &\bf13.03 &16.12  \\
    ESRL-Entropy  & 10  &\bf34.63      &\bf35.51     &\bf80.13  &5.11  &10.02        &27.39     &\bf28.02  &\bf83.90      &13.82 &\bf15.58  \\  \midrule
    REINFORCE     & 15  &34.41      &35.28          &79.88      &9.91      &22.25     &-          &-     &-    &-  &\textgreater24.00  \\
    MRT           & 15  &-      &-          &-      &-      &\textgreater24.00      &-          &-     &-    &-  &\textgreater24.00  \\
    ESRL-Random   & 15  &33.61      &34.34     &79.12      &6.91         &11.86      &27.22     &27.68     &83.71    &15.53 &17.15  \\
    ESRL-BLEU     & 15  &\bf34.79   &\bf35.63     &80.24      &6.10      &12.53      &\bf27.54  &\bf28.13     &\bf83.98   &\bf15.46 &\bf16.97  \\
    ESRL-Entropy  & 15  &34.68      &35.57     &\bf80.33      &\bf6.03 &\bf11.78     &27.45     &28.07     &83.93    &16.08 &17.72  \\  \midrule
    REINFORCE     & 20  &-      &-          &-      &-      &\textgreater24.00      &-          &-     &-    &-  &\textgreater24.00  \\
    MRT           & 20  &-      &-          &-      &-      &\textgreater24.00      &-          &-     &-    &-  &\textgreater24.00  \\
    ESRL-Random   & 20  &33.78    &34.51     &79.34    &7.25    &17.16      &27.18     &27.65     &83.68    &\bf20.13 &21.09  \\
    ESRL-BLEU     & 20  &\bf34.95 &\bf35.82  &\bf80.56 &\bf7.04 &\bf16.35   &\bf27.67  &\bf28.30  &\bf84.12   &21.36 &\bf19.45  \\
    ESRL-Entropy  & 20  &34.83    &35.74     &80.42    &7.32    &16.98      &27.58     &28.21     &84.05    &21.65  &20.76  \\ 
    \bottomrule[1.1pt]
    \end{tabular}}
    \caption{
        Results on the machine translation task using different sampling sizes.
        The best results for each group of student models are in \textbf{bold}.
        The suffix “-Random”, “-BLEU”, and “-Entropy” denote that we use random-based, BLEU-based, and entropy-based strategies to adjust the sampling size, respectively.
        \textbf{SS}: sampling size; \textbf{SBLEU}: \textit{sacre}BLEU score; \textbf{Time}: training time; \textbf{Memory}: maximum memory consumption.
    }
    \label{tab_main_experiment_translation}
    \vspace{-5mm}
\end{table*}

\paragraph{Setups}
For machine translation and abstractive summarization tasks, we pre-trained a standard Transformer base model \cite{vaswani2017attention} using the MLE until convergence.
Here we employed BLEU and ROUGE-L as the reward functions during RL training.
For RLHF, we fine-tuned a LLaMA-7B model using LoRA approach.
Following \citet{ouyang2022training}'s work, we trained a reward model using a LLaMA-7B model to predict rewards during RL training.
More training setups are shown in \textbf{Appendix A}.

\paragraph{Evaluation Metrics}
\label{sec-eval-metrics}
We measured the translation quality in terms of BLEU. 
Both tokenized BLEU and \textit{sacre}BLEU \cite{post2018call} scores were reported on the IWSLT and WMT datasets.
We measured the summary quality by calculating ROUGE-L scores for the CNN/DM dataset.
To further evaluate the performance of the model, two model-based metrics, COMET-22 \cite{rei2022comet} and BARTScore \cite{yuan2021bartscore}, were employed for measuring machine translation and summarization tasks, respectively.
Additionally, we used Vicuna benchmark\footnote{\url{https://lmsys.org/blog/2023-03-30-vicuna/}} to evaluate the performance of RLHF, where the scores were assessed by GPT-4 following \citet{zheng2023judging}'s work.
For training efficiency and memory consumption, we tested our proposed ESRL on four TITAN RTX GPUs.
Specifically, we employed a global batch size (per GPU) of 1,024 tokens, 2048 tokens, and 4 samples for the machine translation, abstractive summarization, and RLHF, respectively.
Note that we also used the restructuring batch trick in MRT baselines to make a fair comparison.

\paragraph{Baselines}
Our baseline is the standard \textbf{MLE}.
Additionally, we compare ESRL with commonly used sampling-based (on-policy) RL methods, including \textbf{REINFORCE} \cite{ranzato2015sequence} and \textbf{MRT} \cite{shen2016minimum}, across various sampling sizes.
For REINFORCE, following \citet{kiegeland2021revisiting}, we implemented it using the moving average baseline with the temperature $\tau=0.95$.
In RLHF, we compare ESRL with the standard \textbf{SFT} and \textbf{PPO}.
We also chose \textbf{ESRL-Random} method as an additional baseline to evaluate the effectiveness of ESRL.
In ESRL-Random, we randomly adjusted the size and temperature of sampling during dynamic sampling.
Furthermore, we compare with off-policy RL methods, including \textbf{GOLD-$s$} and \textbf{GOLD-$p$} \cite{pang2020text}, as shown in Table \ref{tab_comparison_gold}.

\subsection{Experimental Results}

\paragraph{Results of Machine Translation}
Figure \ref{tab_main_experiment_translation} summarizes the results of machine translation.
In terms of training time and memory consumption, our ESRL consistently outperforms REINFORCE and MRT on different sampling sizes.
For instance, ESRL can reduce about 47\% of memory consumption and 39\% of training time on training IWSLT model with a sampling size of 15. 
It demonstrates that ESRL can efficiently achieve RL training on the machine translation task, while also showing its ability to conduct larger-scale sampling with identical settings on resource-constrained devices.
In terms of translation quality, ESRL achieves the best result in training translation models compared to all the baselines.
Notably, ESRL yields a +0.98 BLEU improvement on the WMT En-De dataset compared to MLE, when using the sampling size of 20.
Compared to REINFORCE and MRT, our ESRL can also gain a better translation quality.
For instance, ESRL outperforms MRT by 0.32 SBLEU points on the IWSLT dataset when using the sampling size of 10.
We attribute this to the fact that ESRL benefits from the appropriate exploration obtained by the dynamic sampling at each training step (see an analysis from Section \textbf{Balancing Exploration and Exploitation}).
Additionally, as the comparison of the “BLEU” and “COMET-22” columns in Table \ref{tab_main_experiment_translation}, we observe that the same phenomenon with SBLEU that ESRL can also outperform all baselines over a large margin.  

\begin{table}[]
    \centering
    \scalebox{0.88}{
    \begin{tabular}{lcccrr}
    \toprule[1.1pt]
    Method       & SS & RG-L & BS & \begin{tabular}[c]{@{}c@{}}Time\\ (hours)\end{tabular} & \begin{tabular}[c]{@{}c@{}}Memory\\ (G)\end{tabular} \\ \midrule
    MLE          & -  & 37.06     & -1.65      & -    & - \\ \midrule
    REINFORCE    & 1  & 37.58     & -1.54      &4.38      &9.75   \\
    ESRL-Random  & 1  & 37.23     & -1.63      &2.52      &4.14                                                \\
    ESRL-ROUGE   & 1  & \bf37.72  & \bf-1.48     &\bf2.46     &4.26    \\
    ESRL-Entropy & 1  & 37.64     & -1.50      &2.58      &\bf4.01   \\ \midrule
    REINFORCE    & 5  &-     &-      &-  &\textgreater24.00 \\
    MRT          & 5  &-     &-      &-  &\textgreater24.00 \\
    ESRL-Random  & 5  & 37.38     & -1.61      &4.05      &6.72 \\
    ESRL-ROUGE   & 5  &\bf38.13   & \bf-1.42     &\bf3.87   &\bf6.59     \\
    ESRL-Entropy & 5  &37.98      &-1.46      &4.28      &7.02                    \\ \bottomrule[1.1pt]                      
    \end{tabular}}
    \caption{
        Results on the abstractive summarization task.
        \textbf{RG-L}: ROUGE-L;
        \textbf{BS}: BARTScore.
    }
    \vspace{-6mm}
    \label{tab_main_experiment_summarization}
\end{table}

\paragraph{Results of Abstractive Summarization}
We also evaluated the proposed ESRL on the abstractive summarization task.
The results are presented in Table \ref{tab_main_experiment_summarization}.
We can see that ESRL outperforms MLE by a large margin (\textit{e.g.,} 1.07 ROUGE-L and 0.23 BARTScore benefits).
Due to the excessively long input (\textit{i.e.,} an article), REINFORCE and MRT necessitate a huge training footprint to train a summarization model while sampling multiple sequences.
However, in this case, ESRL still achieves an efficient RL training as the sampling process receives benefits from both two-stage sampling and dynamic sampling approaches.

\paragraph{Results of RLHF}
As shown in Table \ref{tab_rlhf}, we evaluated our ESRL in RLHF with a sampling size of 1.
The experimental results indicate that compared to the conventional PPO, our ESRL can still be more memory-efficient and faster in RLHF.
Notably, ESRL-Entropy can outperform SFT by a substantial margin of 56.00 points on Vicuna's total score.
Additionally, compared to the PPO with two-stage sampling, our ESRL yields a +30.00 points improvement.
It demonstrates that our dynamic sampling approach not only improves the training efficiency but also contributes to the generation quality in RLHF.

Furthermore, compared to ESRL-Random, we observe that ESRL-BLEU, ESRL-Reward, and ESRL-Entropy can achieve better generation quality on all tasks.
It demonstrates that model capacity-based adjustment is superior to random-based adjustment.
Additionally, we investigate the performance gain of different capacity estimation strategies.
From the results, we find that both the entropy-based estimation and the BLEU/ROUGE/Reward-based estimation can contribute to the generation quality improvement over the baselines on all tasks.

\begin{table}[]
    \centering
    \scalebox{0.85}{
    \begin{tabular}{lccc}
    \toprule[1.1pt]
    Method & Vicuna Score & Time (hour) & Memory (G) \\ \midrule
    SFT          &560.00 &- &- \\ \midrule
    PPO          &- &- &\textgreater24.00 \\
    PPO w/ TS    &596.00 &13.87 &23.14   \\
    ESRL-Random  &571.00 &10.81 &19.57 \\
    ESRL-Reward  &619.00 &10.53 &\textbf{19.36} \\
    ESRL-Entropy &\textbf{626.00} &\textbf{10.19} &20.03 \\
    \bottomrule[1.1pt]
    
    \end{tabular}}
    \caption{
        Vicuna's total score assessed by GPT-4.
        “-Reward” denotes that we use the predicted reward score to estimate model capability.
        \textbf{TS}: two-stage sampling.
    }
    \label{tab_rlhf}
\end{table}

\subsection{Ablation Study}
In this section, we present detailed ablation studies to explore effects of two-stage sampling, dynamic sampling, and FIFO-based baseline reward.
The experiments are conducted on the IWSLT dataset, and the impacts of removing each approach are thoroughly examined.
The results are summarized in Table \ref{tab_ablation_study}.
From the results, we see that the two-stage sampling approach can significantly reduce training time cost and memory consumption, which makes it feasible to RL training on resource-constrained devices.
We also see that without dynamic sampling, ESRL fails to gain a well-performed translation model.
Furthermore, to investigate the impact of temperature adjustment, we attempt to employ ESRL to train a translation model with removing this factor, specifically by solely adjusting the sampling size during the dynamic sampling process. 
The results show that temperature adjustment can improve generation quality without bringing additional computational costs.
Additionally, we see that using the FIFO-based baseline reward can train a better model.
It shows the effectiveness of using FIFO to compute baseline value.

\begin{table}[]
    \centering
    \scalebox{0.81}{
    \begin{tabular}{lccccc}
    \toprule[1.1pt]
    Method & SBLEU & BLEU & COMET-22 & \begin{tabular}[c]{@{}c@{}}Time\\ (hours)\end{tabular} & \begin{tabular}[c]{@{}c@{}}Memory\\ (G)\end{tabular} \\ \midrule
    MLE             &35.30 & 36.04  &80.88 &- &- \\ \midrule
    ESRL            &36.34 &37.15  &82.29 &7.04 &16.35   \\ \midrule
    w/o TS          &-   &-  &-    &-  &\textgreater24.00                                                      \\
    w/o DS          &35.87 &36.53  &81.27 &8.89 &19.10                                 \\
    w/o TA          &36.02 &36.76  &81.55 &7.13 &16.42 \\ 
    w/o FBR         &36.17 &36.96  &82.01 &7.02 &15.87 \\ \bottomrule[1.1pt]
    \end{tabular}}
    \caption{
        Ablation studies on the components of ESRL.
        The translation quality is tested on the IWSLT development set.
        \textbf{DS}: dynamic sampling;
        \textbf{TA}: temperature adjustment;
        \textbf{FBR}: FIFO-based baseline reward.
    }
    \vspace{-4mm}
    \label{tab_ablation_study}
\end{table}

\definecolor{mygreen}{RGB}{46,139,87}
\definecolor{myred}{RGB}{238,44,44}
\definecolor{myblue}{RGB}{30,144,255}
\begin{figure}[t!]
\centering
\tikzstyle{every node}=[scale=0.88]
\begin{tikzpicture}
    \scriptsize{
    \begin{axis}
    [
	  at={(0,0)},
      ymajorgrids,
      xmajorgrids,
      grid style=dashed,
      width=.50\textwidth,
      height=.20\textwidth,
      legend style={at={(0.1,0.55)}, anchor=south west},
      legend cell align={left},
      ylabel={\scriptsize{SBLEU}},
      ylabel style={yshift=-4ex, xshift=0em, scale=1.2},,
      legend style={yshift=10pt,xshift=-2.6em, legend plot pos=right,font={\tiny},cells={anchor= west}},
	  legend style={at={(0.1,0.1)},anchor=south west},
	  xmin=0,
	  xmax=0.5,
	  ymax=36.2,
	  ymin=35.2
      ]
      \addplot[blue!60,mark=pentagon*,mark size=1.5pt,thick,mark options={fill=white,draw=blue,line width=0.5pt}] file {images/data/data1.dat};
      \addlegendentry{\scalebox{1.0}{ESRL-BLEU}}

      \addplot[orange!80,mark=triangle*,,mark size=1.5pt,thick,mark options={fill=white,draw=orange,line width=0.5pt}] file {images/data/data2.dat};
      \addlegendentry{\scalebox{1.0}{MLE}}
    \end{axis}}
	\node [anchor=center] at (.2\textwidth,-6ex) {\scalebox{1.2}{(a) Performance Comparison}};
	\scriptsize{
		\begin{axis}
			[ymajorgrids,
			xmajorgrids,
			grid style=dashed,
			anchor=north west,
			at={(0,-5em)},
			width=.26\textwidth,
			symbolic x coords={0,0.1,0.2,0.3,0.4,0.5},
	        xmax=0.5,
	        xmin=0,
			ymin=5.5,
			ymax=9.5,
			xtick=data,
			x tick label style={/pgf/number format/fixed,
				/pgf/number format/fixed zerofill,
				/pgf/number format/precision=1},
			y tick label style={/pgf/number format/fixed,
				/pgf/number format/fixed zerofill,
				/pgf/number format/precision=1},
			ylabel=\footnotesize{\scriptsize Training Time (hours)},
			ylabel style={yshift=-1.6em, scale=1.2},
			legend style={at={(0.5,-0.20)},
				anchor=north,legend columns=-1},]
			\addplot [myred!80,mark=diamond*,line width=.5pt] file {images/data/data3.dat};
		\end{axis}
		\begin{axis}
			[ymajorgrids,
			xmajorgrids,
			grid style=dashed,
			anchor=north west,
			at={(.24\textwidth,-5em)},
			width=.26\textwidth,
			symbolic x coords={0,0.1,0.2,0.3,0.4,0.5},
	        xmax=0.5,
			xmin=0,
			ymin=12.0,
			ymax=20.0,
			xtick=data,
			x tick label style={/pgf/number format/fixed,
				/pgf/number format/fixed zerofill,
				/pgf/number format/precision=1},
			y tick label style={/pgf/number format/fixed,
				/pgf/number format/fixed zerofill,
				/pgf/number format/precision=1},
			ylabel=\footnotesize{\scriptsize Memory (G)},
			ylabel style={yshift=-1.6em, scale=1.2},]
			\addplot [myred!80,mark=diamond*,line width=.5pt] file {images/data/data4.dat};

	\end{axis}}
	\node [anchor=center] at (.2\textwidth,-38ex) {\scalebox{1.2}{(b) Efficiency Comparison}};
	\end{tikzpicture}

	\caption{
		The comparison of performance and efficiency against different elimination ratios: 0, 0.1, 0.2, 0.3, 0.4, 0.5. 
	}
        \vspace{-3mm}
	\label{fig_diff_eli_ratios}
\end{figure}

\subsection{Analysis}
\paragraph{Performance on Different Elimination Ratios}
Based on the two-stage sampling with a sampling size of 20, we investigate the impact of using different elimination ratios.
Figure \ref{fig_diff_eli_ratios} (top) compares ESRL-BLEU with MLE on the IWSLT dataset.
We see that ESRL can achieve consistent SBLEU improvements across various elimination ratios.
Additionally, the results show an interesting observation that the elimination operation may bring certain benefits to our ESRL in terms of SBLEU.
For instance, using an elimination ratio of 0.3 yields an improvement of SBLEU, compared to using an elimination ratio of 0 (\textit{i.e.,} do not use the elimination operation).
We attempt to give a potential cause for this observation from the perspective of balancing exploration and exploitation (See Section \textbf{Balancing Exploration and Exploitation}).
Figure \ref{fig_diff_eli_ratios} (bottom) shows the results for training time and memory consumption using different elimination ratios.
From the results, we can observe that our elimination operation can progressively diminish training time and memory consumption usage as increasing the elimination ratios.
Considering the impact on BLEU and efficiency, we choose the elimination ratio of 0.3 to conduct our all experiments.

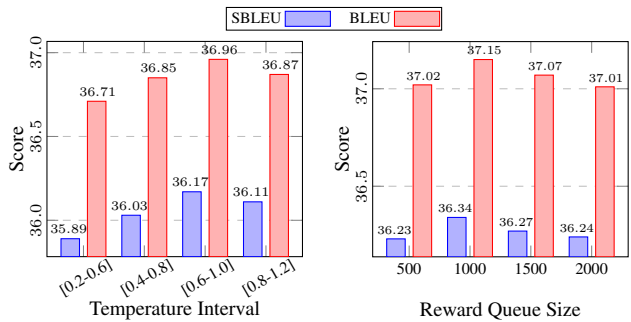
\begin{figure}[t]
    \setlength{\abovecaptionskip}{0.1cm}
    \centering
    \tikzstyle{every node}=[scale=1.1]
    \begin{tikzpicture}
    \centering
      \scriptsize{
      \begin{axis}[
        at={(-2em,0)},
        anchor=south west,
        ymajorgrids,
        grid style=dashed,
        legend style={at={(0.02,0.65)}, anchor=south west},
        legend cell align={left},
        ybar=3pt,
        enlarge x limits=0.5,
        xtick align=inside,
        height=.25\textwidth,
        width=.28\textwidth,
        bar width=1.0em,
        xlabel={Temperature Interval},
        ylabel={Score},
        symbolic x coords={{1}, {2}, {3}, {4}},
        xtick=data,
        nodes near coords align={vertical},
        xticklabels={{[0.2-0.6]},{[0.4-0.8]},{[0.6-1.0]},{[0.8-1.2]}},
        x tick label style={
             rotate=30,
             anchor=center,
             scale=0.8,
             xshift=-.5em,
             yshift=-1.2em
         },
        enlarge x limits=0.2,
        ylabel style={yshift=-3.5em},xlabel style={yshift=0.3em,align=center},
        yticklabel style={/pgf/number format/fixed,/pgf/number format/fixed zerofill,/pgf/number format/precision=1,rotate=90,scale=0.80},
        legend entries={\small SBLEU, \small BLEU},
        legend columns=-1,
        legend style={
                minimum height=2ex,
                inner sep=.6pt,
                at={(0.7, 1.05)},
                nodes={scale=0.6},
                legend cell align=left,
			    legend plot pos=right,
                /tikz/every even column/.append style={column sep=0.20cm}},
        nodes near coords,
        nodes near coords style={font=\tiny, scale=0.75},
        nodes near coords style={/pgf/number format/.cd, fixed zerofill, precision=2},
        ]
            \addplot[fill=blue!30, draw=blue, area legend] coordinates {({1},35.89) ({2},36.03) ({3},36.17) ({4},36.11)};
            \addplot[fill=red!30, draw=red, area legend] coordinates {({1},36.71) ({2},36.85) ({3},36.96) ({4},36.87) };
  
      \end{axis}
      }
  
      \scriptsize{
      \begin{axis}[
        at={(15.6em,0)},
        ymajorgrids,
        grid style=dashed,
        legend style={at={(0.41,0.54)}, anchor=south west},
        legend cell align={left},
        ybar=3pt,
        enlarge x limits=0.5,
        xtick align=inside,
        height=.25\textwidth,
        width=.28\textwidth,
        bar width=1.0em,
        xlabel={Reward Queue Size},
        ylabel={Score},
        symbolic x coords={{1}, {2}, {3}, {4}},
        xtick=data,
        nodes near coords align={vertical},
        xticklabels={500, 1000, 1500, 2000},
        x tick label style={
             scale=0.8,
         },
        enlarge x limits=0.2,
        ylabel style={yshift=-3.5em},xlabel style={yshift=0.3em,align=center},
        yticklabel style={/pgf/number format/fixed,/pgf/number format/fixed zerofill,/pgf/number format/precision=1,rotate=90, scale=0.80},
        nodes near coords,
        nodes near coords style={font=\tiny, scale=0.7},
        nodes near coords style={/pgf/number format/.cd, fixed zerofill, precision=2},
        ]
        \addplot[fill=blue!30, draw=blue, area legend] coordinates {({1},36.23) ({2},36.34) ({3},36.27) ({4},36.24) };
        \addplot[fill=red!30, draw=red, area legend] coordinates {({1},37.02) ({2},37.15) ({3},37.07) ({4},37.01)};
  
      \end{axis}
      }
  \end{tikzpicture}
  \caption{
        Performance of ESRL with different temperature intervals and reward queue sizes on the IWSLT dataset.
    }
  \vspace{-4mm}
  \label{fig_diff_temp_queue_length}
  \end{figure}

\paragraph{Effect of Temperature Interval on Performance}
We conduct experiments to study the impact of using different temperature intervals.
As shown in Figure \ref{fig_diff_temp_queue_length} (left), we swept over different intervals: $\lbrace \left[0.2, 0.6\right], \left[0.4, 0.8\right]$, $\left[0.6, 1.0\right], \left[0.8, 1.2\right] \rbrace $.
From the results, we see that the use of different temperature intervals can result in different performance gains.
We find that the optimal temperature interval is $\left[0.6, 1.0\right]$ which makes an appropriate diversity in the sampled sequences.

\paragraph{Effect of Global Queue Size on Performance}
We also analyze how the global queue size affects our ESRL's performance.
Figure \ref{fig_diff_temp_queue_length} (right) shows the experimental results with different global queue sizes.
We can see that both smaller and larger queue size hurts the BLEU score.
This can be attributed to the fact that a smaller queue size fails to accurately approximate an average of global-level rewards, while a larger global size may contain an excessive number of outdated rewards.

\begin{table}[]
    \centering
    \scalebox{0.74}{
        \begin{tabular}{l|c|ccc}
        \toprule[1.1pt]
        Method        &RL Type      & SBLEU  & BLEU & COMET-22 \\ \midrule
        MLE           &-            &33.77   &34.57  &79.32   \\ 
        ESRL          &On-policy    &34.95   &35.82  &80.56   \\ \midrule
        GOLD-$p$      &Off-policy   &34.21   &35.30  &79.83   \\
        GOLD-$p^*$    &Off-policy   &-       &35.22 &-         \\
        GOLD-$s$      &Off-policy   &34.33   &35.41  &80.11   \\
        GOLD-$s^*$    &Off-policy   &-       &35.45 &-   \\
        ESRL+GOLD-$s$ &On-policy+Off-policy &\bf35.12   &\bf36.05  &\bf80.82    \\
        \bottomrule[1.1pt]
        \end{tabular}
    }
    \caption{
        Performance on the IWSLT test set, using standard models trained with off-policy objectives.
        Starred (*) results are taken from \citet{pang2020text}.
    }
    \vspace{-4mm}
\label{tab_comparison_gold}
\end{table}

\paragraph{Comparison with Off-policy RL Methods}
Table \ref{tab_comparison_gold} shows the performance of off-policy RL method on the IWSLT dataset.
We can observe that ESRL is still better than strong GOLD \cite{pang2020text} under the evaluation of various metrics.
Furthermore, our ESRL is orthogonal to the off-policy RL method.
Here, we take GOLD-$s$ as an instance.
Specifically, we first train a translation model with ESRL, and then use the trained model to perform GOLD-$s$ procedure.
The experimental results show that the combined method can achieve superior performance.

\paragraph{Balancing Exploration and Exploitation}
\label{secbal}
Balancing exploration and exploitation has been proved to improve RL in the planning problem \cite{tokic2010adaptive, sutton2018reinforcement, jiang2020generative}. 
Here, we attempt to illustrate the observation that our ESRL can achieve better performance than all baselines from a perspective of effectively achieving the exploration-exploitation balance.
When the model has a strong capacity and obtains high deterministic rewards, our ESRL exploits and reduces exploration as much as possible, \textit{i.e.,} reducing the size and temperature of sampling.
This allows the model to make full use of the current learned knowledge for decision-making and optimization.
Instead, when the model has a weak capacity, ESRL increases the size and temperature of sampling to enhance exploration, which gathers more possible generated sequences to optimize the model.
Thus, compared to baselines, using dynamic sampling approach enables ESRL to balance exploration and exploitation well and achieve better performance.

See more analysis in \textbf{Appendix B}.

\section{Conclusion}
In this paper, we focus on reducing the computational cost of RL training in sequence generation models.
We have proposed an efficient sampling-based RL method (referred to as ESRL) via two-stage sampling and dynamic sampling approaches.
Our extensive experiments show that our ESRL significantly outperforms all baselines in terms of both training efficiency and generation quality.

\appendix
\nobibliography*

\bibliography{aaai23}

\clearpage

\section{Appendix A: Experimental Details}
\paragraph{Dateset Preprocessing}
We conducted experiments on four machine translation tasks: IWSLT'14 De-En, En-De, WMT'18 De-En, and En-De.
For IWSLT'14 tasks, we lowercased all the sentences.
For WMT'18 tasks, we followed the same dataset setting as \citet{yehudai2022reinforcement}.
Here, we adopted a joint source and target BPE factorization with vocabulary sizes of 10K and 32K for IWSLT'14 and WMT'18 tasks, respectively.
In addition, we used scripts from the Moses toolkit\footnote{\url{https://github.com/moses-smt/mosesdecoder/tree/master/scripts}} to preprocess all datasets, including normalize punctuation and tokenization.
We filtered out the sentences longer than 250 tokens, and the word ratio between the source and the target exceed 1:1.5 or 1.5:1.

\paragraph{Datasets Statistics}
The statistical information on the utilized datasets is summarized in Tables \ref{tab_statistical_mt_summay_datasets} and \ref{tab_statistical_rlhf_datasets}.

\begin{table}[h]
    \centering
    \scalebox{0.80}{
        \begin{tabular}{l|ccc}
        \toprule[1.1pt]
        -             & IWSLT’14 De-En & WMT’14 En-De & CNN/DM \\ \midrule
        Train         &160,239      &3,896,364  &287,113            \\ 
        Valid         &7,283      &39,388   &13,368        \\
        Test          &6,750      &3,003   &11,490          \\ \bottomrule[1.1pt]
        \end{tabular}
    }
    \caption{
        Statistical information on the machine translation and abstractive summarization datasets.
    }
\label{tab_statistical_mt_summay_datasets}
\end{table}

\begin{table}[h]
    \centering
    \scalebox{0.80}{
        \begin{tabular}{l|c}
        \toprule[1.1pt]
        -                       &Sample Size  \\ \midrule
        Alpaca’s 52k            &52,002      \\ 
        GPT-4 Alpaca (English)  &52,002        \\
        GPT-4 Comparison (English) &36,441 \\ 
        \bottomrule[1.1pt]
        \end{tabular}
    }
    \caption{
        Statistical information on the RLHF datasets.
        GPT-4 Alpaca data is created by using the Alpaca dataset as inputs, and replacing the example generations with generations from GPT-4.
    }
\label{tab_statistical_rlhf_datasets}
\end{table}

\paragraph{Setups}
In the process of RL training on machine translation and abstractive summarization tasks, we initialized the model using MLE checkpoint with the highest score (BLEU and ROUGE-L) on the development set.
For the IWSLT’14 De-En and WMT’14 En-De datasets, we trained the pre-trained model for 10 epochs and 8000 steps with a batch size of 4096 tokens (token level), respectively.
For the summarization task, we trained the pre-trained model for about 8000 steps with a batch size of 4096 tokens.
For fair comparisons, we re-implemented all baselines with our strong pre-trained model under the same training settings as \citet{kiegeland2021revisiting}.
Moreover, we used the learning rate of 3e-5 for all tasks.
In ESRL, $\beta$, $\tau_{min}$, $\tau_{max}$, and $\mathcal{Q}_{size}$ were set to 0.1, 0.6, 1.0, and 1000, respectively. 
During the sampling process, we employed top-$k$ sampling, where $k$ was set to 50.
In RLHF, we fine-tuned a LLaMA-7B model using LoRA approach.
During training reward model, we set training epoch, batch size (sequence level), and learning rate to 1, 4, and 1e-5, respectively.
During RL training, we set the training step, new target length, learning rate, and batch size (sequence level) to 6000, 150, 1e-5, and 4, respectively.
Our codebase is built based on \texttt{Fairseq} \cite{ott2019fairseq} and \texttt{LLaMA Efficient Tuning} \cite{llama-efficient-tuning}, and it would be publicly available soon.

\section{Appendix B: More Analysis}
\begin{table}[]
    \centering
    \scalebox{0.80}{
        \begin{tabular}{l|ccc}
        \toprule[1.1pt]
        Method       & SBLEU & BLEU & COMET-22 \\ \midrule
        MLE                 &35.30      &36.04  &80.88            \\ \midrule
        ESRL+Beam Search         &35.73      &36.46   &81.28       \\
        ESRL+Diverse Beam Search &36.01      &36.78   &81.53          \\
        ESRL+Top-$p$ Sampling    &36.18      &37.01   &81.94  \\
        ESRL+Top-$k$ Sampling    &\bf36.34 &\bf37.15  &\bf82.29                              \\ \bottomrule[1.1pt]
        \end{tabular}
    }
    \caption{
        The comparison of translation quality against different sampling strategies.
    }
\label{tab_other_sampling_methods}
\end{table}
\paragraph{Performance on Different Sampling Strategies}
We conduct experiments employing alternative sampling strategies on the IWSLT dataset, including beam search \cite{freitag2017beam}, diverse beam search \cite{roberts2020decoding}, top-$p$ sampling \cite{holtzman2019curious}, and top-$k$ sampling.
The results shown in Table \ref{tab_other_sampling_methods} indicate that top-$k$ sampling can yield the best performance for training a sequence model.
We conjecture that top-$k$ sampling has the advantage to generate sequences that have more diversity while maintaining a high quality compared to other methods, which makes it better suited for our ESRL.

\paragraph{Details of Sampling}
For an input $x$, we sample a candidate generation sequence $\hat{y}=\{ \hat{y}_{1}, \hat{y}_{2}, \cdots, \hat{y}_{N}\}$ with the model by an autoregressive mode.
Specifically, at time step $t$, we use the Monte Carlo method to sample a token from the multinomial distribution based on the previously sampled tokens $\hat{y}_{<t}$.
In general, this multinomial distribution is given by output logits $z=[z_{1}, z_{2}, \cdots, z_{|\mathcal{V}|}]$ of the model through a temperature factor $\tau$.
The formation is as follows:
\begin{equation}
	\scalemath{1.0}{
	\begin{aligned}
		\mathcal{Z}_\theta(z|\hat{y}_{<t},x,\tau)
		& = z/\tau -  \log(\sum_{i=1}^{|\mathcal{V}|} e^{z_i/\tau}) \\
		& = \log(e^{z/\tau}) - \log(\sum_{i=1}^{|\mathcal{V}|}e^{z_i/\tau})     \\
		& = \log(\frac{e^{z/\tau}}{\sum_{i=1}^{|\mathcal{V}|} e^{z_i/\tau}})
	\end{aligned}
	}
\end{equation}
where $\mathcal{Z}_\theta(z|\hat{y}_{<t},x,\tau)$ is the corresponding multinomial distribution at the $t$-th time step and $|\mathcal{V}|$ is the size of target vocabulary.
In practice, it is equivalent to performing log derivative and softmax operations on output logits.
In this case, the temperature factor controls the relative differences of the distribution $\mathcal{Z}_{\theta}$.
A decrease in the temperature factor makes this distribution peakier, thereby leading to a more deterministic sampling process, characterized by a decrease in the diversity of sampled sequences.
Conversely, an increase in the temperature factor leads the distribution to be smoother, which makes a less deterministic sampling process, with an increase in the diversity of sampled sequences.

\end{document}